\documentclass[preprint,12pt]{elsarticle}

\usepackage{amssymb}
\biboptions{sort&compress}
\usepackage{subfigure}
\usepackage{graphicx}%
\usepackage{multirow}%
\usepackage{amsmath,amssymb,amsfonts}%
\usepackage{amsthm}%
\usepackage{mathrsfs}%
\usepackage{xcolor}%
\usepackage{textcomp}%
\usepackage{manyfoot}%
\usepackage{booktabs}%
\usepackage{algorithm}%
\usepackage{algorithmicx}%
\usepackage{algpseudocode}%
\usepackage{listings}%
\usepackage{url}%
\usepackage{hyperref}
\usepackage{amsmath}
\usepackage{booktabs}
\usepackage{array}
\usepackage{booktabs}

\journal{Elsevier}

\begin{document}

\begin{frontmatter}



\title{QGAN-based data augmentation for hybrid quantum-classical neural networks}

\author[label1,label2,label3]{Run-Ze He}
\author[label1]{Jun-Jian Su}
\author[label1]{Su-Juan Qin \corref{cor1}}
\author[label1]{Zheng-Ping Jin \corref{cor1}}
\author[label1]{Fei Gao}
\cortext[cor1]{Corresponding author, E-mail:qsujuan@bupt.edu.cn; zhpjin@bupt.edu.cn}

\affiliation[label1]{organization={Networking and Switching Technology, Beijing University of Posts and Telecommunications},
            city={Beijing},
            postcode={100876}, 
            country={China}}

\affiliation[label2]{organization={National Engineering Research Center of Disaster Backup and Recovery, Beijing University of Posts and Telecommunications},
            city={Beijing},
            postcode={100876}, 
            country={China}}

\affiliation[label3]{organization={School of Cyberspace Security, Beijing University of Posts and Telecommunications, Beijing University of Posts and Telecommunications},
            city={Beijing},
            postcode={100876}, 
            country={China}}

\begin{abstract}
Quantum neural networks converge faster and achieve higher accuracy than classical models. However, data augmentation in quantum machine learning remains underexplored. To tackle data scarcity, we integrate quantum generative adversarial networks (QGANs) with hybrid quantum-classical neural networks (HQCNNs) to develop an augmentation framework. We propose two strategies: a general approach to enhance data processing and classification across HQCNNs, and a customized strategy that dynamically generates samples tailored to the HQCNN's performance on specific data categories, improving its ability to learn from complex datasets. Simulation experiments on the MNIST dataset demonstrate that QGAN outperforms traditional data augmentation methods and classical GANs. Compared to baseline DCGAN, QGAN achieves comparable performance with half the parameters, balancing efficiency and effectiveness. This suggests that QGANs can simplify models and generate high-quality data, enhancing HQCNN accuracy and performance. These findings pave the way for applying quantum data augmentation techniques in machine learning. 
\end{abstract}


\end{frontmatter}



\section{Introduction}\label{sec1}

Quantum computing revolutionizes computational science by using quantum principles such as superposition, entanglement, and interference to solve problems beyond classical systems \cite{bib1}. This has given rise to Quantum Machine Learning (QML), which combines quantum computing with artificial intelligence to enhance learning algorithm efficiency \cite{bib2}. QML leverages quantum properties to handle high-dimensional data and complex tasks, offering advantages over classical machine learning in areas like optimization, pattern recognition, and quantum chemistry \cite{bib3,bib4,bib5}. As quantum hardware advances, QML is emerging as a key driver of computational innovation across various fields.

Hybrid quantum-classical neural networks (HQCNNs) integrate the computational efficiency of classical neural networks with the unique capabilities of quantum circuits, such as superposition and entanglement, offering promise in quantum machine learning. Classical components perform feature extraction and preprocessing, preparing data for quantum processing. Quantum circuits use superposition to process multiple states simultaneously and entanglement to capture complex data correlations, improving efficiency. This hybrid architecture outperforms traditional neural networks in tasks such as image classification, financial forecasting, and drug design \cite{bib6,bib7,bib8,bib9,bib10,bib11}. However, data scarcity remains a major challenge for HQCNNs. Like other models, HQCNNs require large, high-quality datasets for robust generalization and to avoid overfitting. Data acquisition in quantum computing faces challenges: preparing quantum states requires advanced equipment and time, while noisy intermediate-scale quantum (NISQ) hardware has limitations such as few qubits and high error rates \cite{bib12,bib13}. Classical data augmentation methods, like geometric transformations, are less effective for HQCNNs due to differences in quantum data processing \cite{bib14,bib15}. Therefore, developing customized data augmentation strategies for HQCNNs is a critical research priority.

Quantum generative adversarial networks (QGANs) offer a promising solution to the data scarcity issue in quantum machine learning (QML). By adapting the generative adversarial network (GAN) framework to the quantum domain, QGANs use parameterized quantum circuits to generate synthetic quantum states or classical data distributions \cite{bib16}. Theoretical studies show that QGANs, utilizing quantum parallelism and superposition, can achieve exponential speedups in specific tasks \cite{bib17,bib18}. Research confirms QGANs' ability to model complex quantum distributions, highlighting the advantages of quantum superposition and entanglement in generative tasks \cite{bib19,bib20,bib21}. QGAN research spans several areas: at the data processing level, QGANs efficiently learn both classical and quantum data \cite{bib22,bib23,bib24}; in algorithmic optimization, quantum entanglement helps address challenges like non-convex optimization and mode collapse seen in classical GANs \cite{bib25}; and in applications, QGANs show potential in fields like small-molecule drug discovery \cite{bib26} and anomaly detection \cite{bib27}. Unlike classical GANs, which struggle with quantum data distributions, QGANs are naturally compatible with quantum environments, making them ideal for augmenting datasets for hybrid quantum-classical neural networks (HQCNNs) \cite{bib28}. Despite significant theoretical advancements and proof-of-concept validations, practical QGAN applications for HQCNN data augmentation remain underdeveloped. Existing studies primarily focus on theoretical model development, with limited progress in scalable, engineering-oriented solutions for real-world applications. Bridging the gap between the theoretical promise of QGANs and their practical implementation as data augmentation strategies remains a critical challenge in quantum machine learning.

This study presents a framework that integrates hybrid quantum-classical neural networks (HQCNNs) with quantum generative adversarial networks (QGANs) to address the data scarcity issue in quantum machine learning. The research designs two augmentation strategies: a general strategy applicable to various HQCNN architectures for performance optimization, and a customized strategy that dynamically generates samples based on the model’s performance on specific data categories, enhancing robustness on challenging classes. Simulation experiments on the MNIST dataset compare QGAN’s effectiveness with traditional data augmentation methods and classical GANs, confirming the framework’s efficacy. Results show QGAN achieves performance similar to deep convolutional GANs (DCGANs) with fewer parameters, highlighting quantum computing's efficiency in generative modeling. These findings provide a new approach to overcoming limitations of conventional data augmentation and lay the foundation for practical quantum data augmentation in real-world applications.

The rest of the framework of this paper is as follows: In Sect. \ref{sec2}, we provide a comprehensive review of the theoretical foundations of qubits, quantum circuits, variational quantum circuits, and QGANs. In Sect. \ref{sec3}, we elaborate on the proposed architecture design of HQCNN and QGAN, alongside detailed technical descriptions of the two data augmentation strategies. In Sect. \ref{sec4}, we present the experimental results, comparing the proposed approach with classical benchmark models and providing a thorough analysis. In Sect. \ref{sec5}, we conclude the study, summarizing key findings, discussing practical implications, and outlining directions for future research.

\section{Preliminary}\label{sec2}

\subsection{Qubits, Quantum Gates, Quantum Circuits and Measurements}\label{subsec1}

Unlike classical bits, which are limited to 0 or 1, quantum bits (qubits) can exist in a superposition state of $\vert 0\rangle$ and $\vert 1\rangle$. Their state can be described as $\vert\psi\rangle=\alpha|0\rangle+\beta|1\rangle$, where $\alpha, \beta$ are complex coefficients satisfying $\mid\alpha\mid^2+\mid\beta\mid^2=1$. This superposition enables quantum computing to process multiple states simultaneously.

Quantum gates, the fundamental units of quantum computation, operate on qubits in superposition, analogous to classical logic gates. Single quantum bit gates, including the $I$, Pauli-$X$, Pauli-$Y$, Pauli-$Z$, and Hadamard gates, transform the state of a single qubit, as shown in Eq. \ref{it:1}. Phase (rotation) gates are another common class of quantum gates. They keep $\vert 0\rangle$ unchanged, but rotate $\vert 1\rangle$ around the $Z$-axis by a specified angle $\theta$. Commonly used rotation gates in quantum computing and quantum machine learning are mainly ${R}_{x}, {R}_{y}, {R}_{z}$ gates, and the matrix form is shown in Eq. \ref{it:2}. 
Multi-qubit gates, like the controlled-NOT (CNOT) gate, create entanglement and enable multi-qubit operations, with their matrix form given in Eq. \ref{it:3}. The combination of these quantum gates constitutes a quantum circuit, and various complex quantum algorithms can be implemented by designing specific gate sequences. 

Measurement is a critical step in quantum computing for extracting results. Upon measurement, a qubit’s superposition collapses to $\vert 0\rangle$ or $\vert 1\rangle$ state with the probability of $\mid\alpha\mid^2$ and $\mid\beta\mid^2$, respectively, irreversibly altering the qubit’s original state—a fundamental feature of quantum mechanics \cite{bib29}. In entangled systems, measuring one qubit instantly influences others—a principle exploited in protocols like quantum key distribution \cite{bib30} and quantum teleportation \cite{bib31}. Measurement not only converts quantum states into classical information but also bridges quantum and classical computation.

\begin{center}\begin{equation}\label{it:1}
\begin{array}{l}
I=\left[\begin{array}{cc}
1 & 0 \\
0 & 1
\end{array}\right], 
X=\left[\begin{array}{cc}
0 & 1 \\
1 & 0
\end{array}\right] \\
Y=\left[\begin{array}{cc}
0 & -i \\
i & 0
\end{array}\right], Z=\left[\begin{array}{cc}
1 & 0 \\
0 & -1
\end{array}\right]\\
H=\frac{1}{\sqrt{2}}\left[\begin{array}{cc}
1 & 1 \\
1 & -1
\end{array}\right]
\end{array}
\end{equation}\end{center}

\begin{center}\begin{equation}\label{it:2}
\begin{array}{l}
R_{\theta}=\left[\begin{array}{cc}
1 & 0 \\
0 & e^{i \theta}
\end{array}\right],
R_x{}=\left[\begin{array}{cc}
\cos \frac{\theta}{2} & -i \sin \frac{\theta}{2} \\
-i \sin \frac{\theta}{2} & \cos \frac{\theta}{2}
\end{array}\right],\\
R_y{}=\left[\begin{array}{cc}
\cos \frac{\theta}{2} & -\sin \frac{\theta}{2} \\
\sin \frac{\theta}{2} & \cos \frac{\theta}{2}
\end{array}\right],
R_z{}=\left[\begin{array}{cc}
e^{-i \theta / 2} & 0 \\
0 & e^{i \theta / 2}
\end{array}\right].
\end{array}
\end{equation}\end{center}

\begin{center}
\begin{equation}\label{it:3}
\text{CNOT}=
\begin{array}{l}
\begin{bmatrix}
1 & 0 & 0 & 0 \\
0 & 1 & 0 & 0 \\
0 & 0 & 0 & 1 \\
0 & 0 & 1 & 0
\end{bmatrix}
\end{array}
\end{equation}
\end{center}

\subsection{Variational Quantum Circuits(VQCs) and HQCNNs}\label{subsec2}

Variational quantum circuits (VQCs), also known as parameterized quantum circuits (PQCs), comprise gates with tunable parameters. These gates are configured in specific combinations tailored to the task and available quantum resources. The parameter optimization process integrates both classical and quantum systems for effective training. These circuits are remarkably resistant to noise, making them ideal for noisy intermediate-scale quantum (NISQ) devices.

\begin{figure*}[htbp]
\centering
\subfigure{
    \begin{minipage}[t]{1\linewidth}
        \centering
        \includegraphics[width=5in]{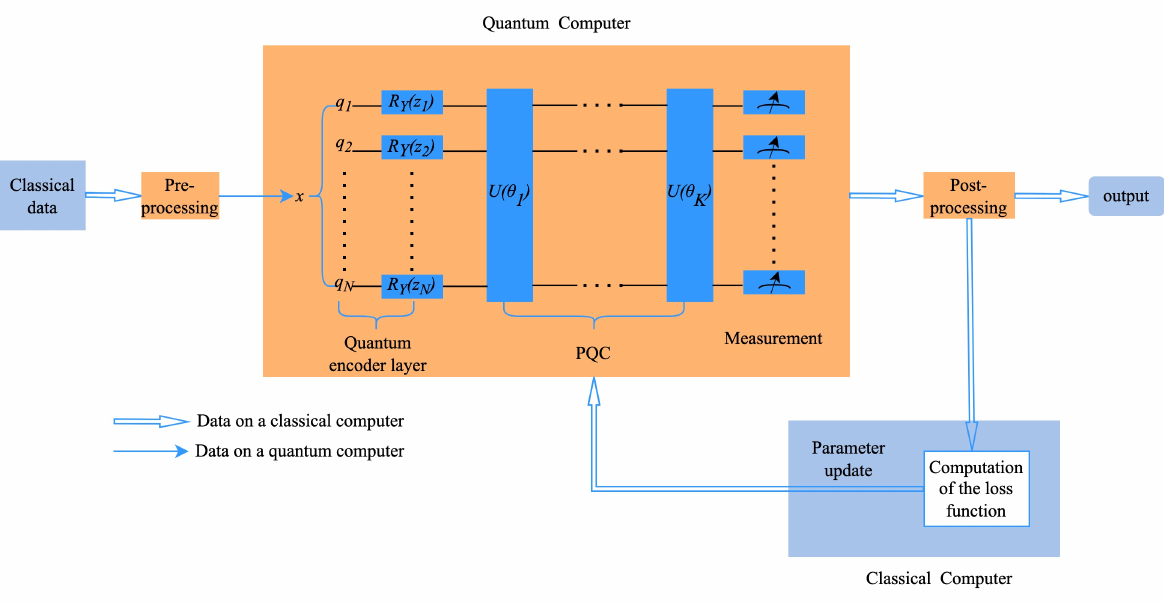}
    \end{minipage}%
}%
\centering
\caption{Flowchart of Hybrid Quantum-Classical Neural Networks Framework}
\label{Fig.1}
\end{figure*}

The workflow of the hybrid quantum-classical neural networks (HQCNNs) reflects the close collaboration between classical and quantum. First, the classical data $\boldsymbol{x}\in\mathbb{R}^d$ is subjected to preliminary feature extraction and processing by the classical neural network module to extract more representative features. Then, these features are mapped to quantum states $\rho(\boldsymbol{x})$ through a specific encoding method, which is achieved by using the quantum circuit $U(\boldsymbol{x})$ to act on the initial state $\rho_0$. Subsequently, a parameterized quantum circuit $V(\boldsymbol{\theta})$ with trainable parameters acts on $\rho(\boldsymbol{x})$ to extract deep features using the superposition and entanglement properties of quantum states, this process combines the powerful parallel computing capabilities of quantum systems and the flexible optimization and adjustment capabilities of classical systems. After that, the resulting state $\rho(\boldsymbol{x})=U(\boldsymbol{x})\rho_0U^\dagger(\boldsymbol{x})$ is measured for a carefully selected observable $O$ to produce an expected value: $E(\boldsymbol{\theta}) = \mathrm{Tr}\left[V(\boldsymbol{\theta}) \rho(\boldsymbol{x}) V^\dagger(\boldsymbol{\theta}) O\right]$. This expected value $E(\boldsymbol{\theta})$ is used to calculate the loss function $L(\boldsymbol{\theta})$, which measures the difference between the HQCNNs predicted output and the expected target. To optimize the model, the parameter $\boldsymbol{\theta}$ is iteratively adjusted on a classical computer according to the loss function to ensure seamless interaction between quantum computing and classical optimization, thereby continuously improving the performance of the model. The complete schematic diagram is shown in Fig. \ref{Fig.1}.

Numerous experiments show that hybrid variational quantum circuits combined with deep neural networks (DNNs) outperform traditional DNNs in efficiency and convergence \cite{bib32,bib33}. Recent studies further show that quantum circuits achieve faster learning and higher accuracy in applications such as quantum convolutional networks \cite{bib34}, quantum LSTM networks \cite{bib35}, and natural language processing \cite{bib36}, underscoring their potential in advancing machine learning.

\subsection{Quantum Generative Adversarial Networks(QGANs)}\label{subsec3}

Generative adversarial networks (GANs), a breakthrough in unsupervised learning, operate based on the principle of adversarial training. GANs perform well in image generation \cite{bib37}, video synthesis, speech generation \cite{bib38}, and text generation \cite{bib39}, with broad applications. GANs were first proposed by Goodfellow et al. in 2014 \cite{bib40}. A GAN consists of two key components: a generator (G) and a discriminator (D). The generator takes random samples from a latent space to produce outputs resembling real training data, while the discriminator learns to distinguish real from generated samples. Through adversarial training, the generator learns to fool the discriminator, and both networks iteratively adjust parameters until the discriminator assigns equal (50\%) probability to real and generated samples, indicating a Nash equilibrium. Quantum GANs (QGANs) also comprise a generator and a discriminator, but the generator uses parameterized quantum circuits instead of classical networks. The core adversarial mechanism remains unchanged. The loss function for training QGANs is defined in Eq. \ref{it:4}. Training alternates between optimizing the generator’s parameters $\lambda$ and the discriminator’s parameters $\mu$, similar to classical GANs. Quantum generator parameters are optimized using automatic differentiation \cite{bib41}. When training the discriminator, $\lambda$ is fixed and $\mu$ is updated using backpropagation. When training the generator, $\mu$ is fixed and gradients are computed using the parameter shift rule \cite{bib42}. The training process is illustrated in Algorithm \ref{Al.1}.

\begin{center}\begin{equation}\label{it:4}
\min_{G}\max_{D}V(D,G)=\mathbb{E}_{\boldsymbol{x}\sim p_{\mathrm{data}}(\boldsymbol{x})}[\log D(\boldsymbol{x})]+\mathbb{E}_{\boldsymbol{z}\sim p_{\boldsymbol{z}}(\boldsymbol{z})}[\log(1-D(G(\boldsymbol{z})))]
\end{equation}\end{center}

\makeatletter
\renewcommand{\algorithmiccomment}[1]{%
  \hfill \texttt{\#} #1%
}
\makeatother

\begin{center}
\begin{algorithm}
\footnotesize
\caption{Quantum Generative Adversarial Network Algorithm}
\label{Al.1}
\textbf{Input:} Generator G: qubit $N$, Depth $d$, Trainable parameters $\lambda$; 
Discriminator D: Classical fully-connected neural network, Trainable parameters $\mu$; Learning rate $\eta$, Training set $T$\par
\textbf{Output:} Parameters $\lambda$ of the converged quantum generator $G_{\lambda}$ and parameters $\mu$ of the classical discriminator $D_{\mu}$
\begin{algorithmic}[1]
\While{$\lambda$, $\mu$ have not converged}
    \State Train the discriminator D
    \State Sample a data point $x \sim \kappa $ from the training set $T$
    \State Generate a sample $x_G \sim p_{\lambda}=\vert\langle x\vert\psi_{\lambda}\rangle\vert^2$ from the quantum circuit.
    \State Compute $\nabla_{\mu}\ell(D_{\mu})$ using the backpropagation algorithm
    \State Update $\mu$: $\mu\leftarrow\mu - \eta\nabla_{\mu}\ell(D_{\mu})$
    \State Train the generator G
    \State Adjust the parameters $\lambda$ of the quantum generator to $\lambda + \pi/2$, and sample $x \sim p_{\lambda + \pi/2}(x)=\vert\langle x\vert\psi_{\lambda + \pi/2}\rangle\vert^2$
    \State Adjust the parameters $\lambda$ of the quantum generator to $\lambda - \pi/2$, and sample $x \sim p_{\lambda - \pi/2}(x)=\vert\langle x\vert\psi_{\lambda - \pi/2}\rangle\vert^2$
    \State Compute $\nabla_{\lambda}\ell(G_{\lambda})$
    \State Update $\lambda$: $\lambda\leftarrow\lambda - \eta\nabla_{\lambda}\ell(G_{\lambda})$
\EndWhile
\State \textbf{Return} Parameters $\lambda$ of the converged quantum generator $G_{\lambda}$ and parameters $\mu$ of the classical discriminator $D_{\mu}$
\end{algorithmic}
\end{algorithm}
\end{center}

\section{Methodology}\label{sec3}

In this study, we deeply explore the hybrid quantum-classical neural network (HQCNN) and quantum generative adversarial network (QGAN), aiming to tap the huge potential of quantum mechanics and quantum machine learning in data augmentation, and focus on the impact of different data augmentation strategies on HQCNN model optimization. The study consists of two parts. The first proposes a general data augmentation strategy that is broadly applicable and compatible with various HQCNN architectures. It enhances data processing and classification performance through standardized procedures; The second part introduces a customized augmentation scheme tailored to a specific HQCNN. This scheme analyzes the performance of the model in different categories and constructs a dynamic sample generation mechanism to address overfitting in data-scarce settings and improve learning on complex distributions. The study highlights the unique advantages of quantum computing in data augmentation and offers innovative approaches for enhancing quantum machine learning performance in practical applications.

\subsection{Model Architecture of HQCNN}\label{subsec1}

\begin{figure*}[htbp]
    \centering
    \begin{minipage}[t]{1\linewidth}
        \centering
        \includegraphics[width=3in]{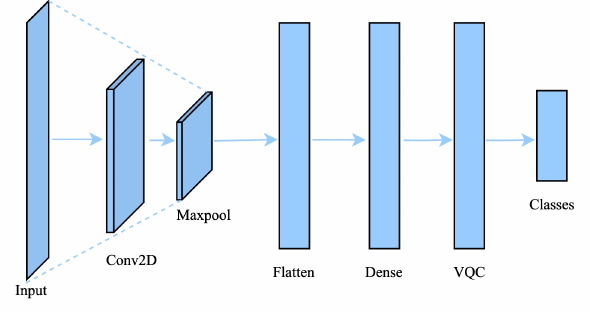}
    \end{minipage}
    \caption{Architecture Diagram of Hybrid Quantum-Classical Neural Network}
    \label{Fig.2}
\end{figure*}

The proposed HQCNN adopts a cascaded architecture combining a classical convolutional neural network (CNN) with a variational quantum circuit (VQC). The CNN module includes a 3×3 convolutional layer, a max-pooling layer, and a ReLU activation. This module effectively extracts local features and reduces the dimensionality of the feature map. The ReLU introduces nonlinearity to enhance model expressiveness. The extracted feature vector is then encoded into a quantum state using angle encoding. This method enables efficient conversion of classical features into quantum states for further quantum processing. The encoded quantum state is fed into a VQC composed of ${R}_{y}$, ${R}_{z}$ rotation gates, and CNOT gates. The full HQCNN architecture is illustrated in Fig. \ref {Fig.2}. This design combines the efficiency of classical CNNs in feature extraction and reduction with the strengths of quantum circuits in deep feature processing and parallelism, offering strong support for classification tasks.

\subsection{Model Architecture of QGAN}\label{subsec1}

\begin{figure*}[htbp]
    \centering
    \begin{minipage}[t]{1\linewidth}
        \centering
        \includegraphics[width=6in]{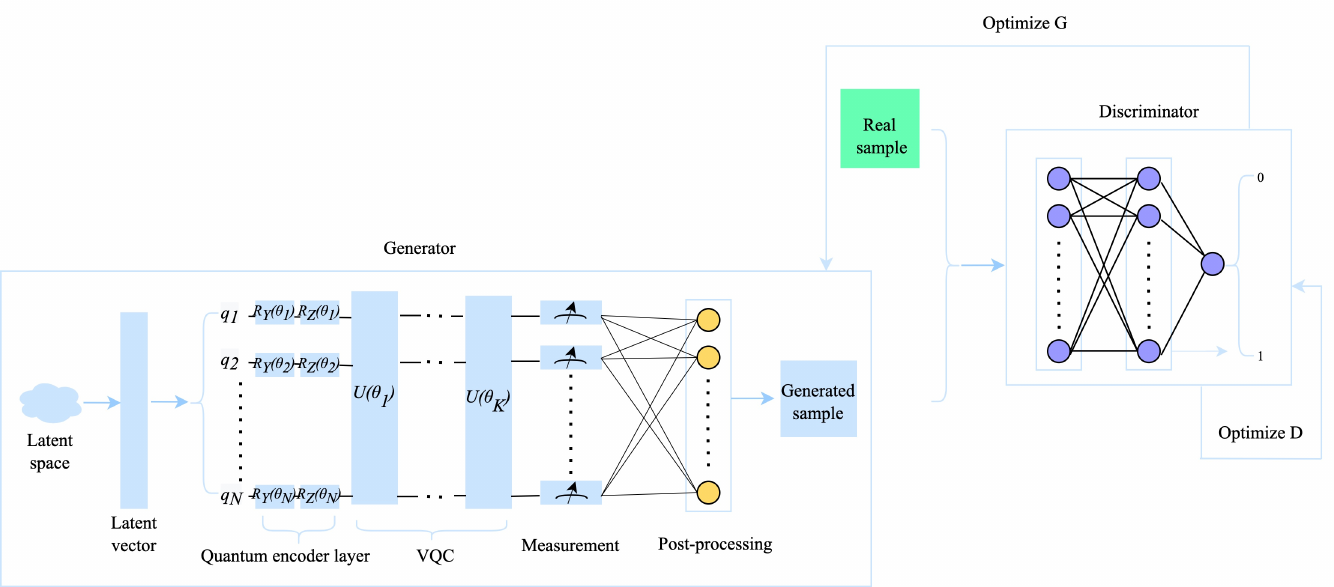}
    \end{minipage}
    \caption{Architecture Diagram of Quantum Generative Adversarial Network}
    \label{Fig.3}
\end{figure*}

Due to the limited quantum resources of Noisy Intermediate-Scale Quantum (NISQ) devices, generating large-scale data directly on quantum hardware is impractical. Moreover, QGANs use random noise vectors as input, requiring lower computational accuracy and offering advantages over other quantum machine learning algorithms. To address these constraints, we adopt a quantum-classical hybrid architecture tailored for QGANs. The generator consists of a quantum variational circuit (VQC) and a classical neural network. The VQC leverages quantum gates and tunable parameters to exploit superposition and entanglement to generate quantum states with data features and complex correlations. The classical neural network post-processes the VQC output, mapping quantum states to data samples aligned with the training distribution via linear transformations and nonlinear activations. The discriminator is a classical neural network with multiple fully connected layers, capable of efficiently distinguishing generated from real samples. The architecture is illustrated in Fig. \ref{Fig.3}.

\subsection{Algorithm of the General-purpose Data Augmentation Strategy}\label{subsec1}

To enhance the performance of HQCNNs, we propose a general data augmentation scheme. This scheme enhances data processing and classification performance through standardized procedures. It uses a quantum generative adversarial network (QGAN) to generate high-quality samples, which are merged with the original dataset to expand data volume and improve generalization. Implementation details are shown in Algorithm \ref{Al.2}.

\makeatletter
\renewcommand{\algorithmiccomment}[1]{%
  \hfill \texttt{\#} #1%
}
\makeatother

\begin{algorithm}
\footnotesize
\caption{General Data Augmentation for HQCNNs}
\label{Al.2}
\textbf{Input:} HQCNN model $M$, training set $D_{train}$, test set $D_{test}$\\
\textbf{Output:} Retrained HQCNN model $M'$
\begin{algorithmic}[1]
\State Initialize QGAN: generator $G$, discriminator $D$
\State Initialize augmented sample set $D_{aug} = []$
\State Train $G$ and $D$ until QGAN converges
\State Generate $N_{gen}$ samples and add to $D_{aug}$
\State Combine datasets: $D_{combined} \gets D_{train} \cup D_{aug}$
\State Retrain $M$ on $D_{combined}$
\State \textbf{Return} $M'$ 
\end{algorithmic}
\end{algorithm}

\subsection{Algorithm of the Customized Data Augmentation Strategy}\label{subsec1}

To enhance a specific HQCNN’s performance in low-accuracy categories without compromising high-accuracy ones, we propose a customized data augmentation method. The method consists of four components: classifier model, sampling module, generation model, and dataset augmentation module. 

Classifier model: computes initial accuracy and confidence for each class and filters generated samples.

Sampling module: estimates the proportion of misclassified samples per class to guide sample generation and allocation.

Generation model: produces samples to expand the training set.

Dataset augmentation module: merges generated and original data to build the enhanced dataset.
The complete process is illustrated in Algorithm \ref{Al.3}.

\makeatletter
\renewcommand{\algorithmiccomment}[1]{%
  \hfill \texttt{\#} #1%
}
\makeatother

\begin{algorithm}
\footnotesize
\caption{Customized Data Augmentation for HQCNNs}
\label{Al.3}
\textbf{Input:} HQCNN model $M$, training set $D_{train}$, test set $D_{test}$, number of classes $C$, basic confidence threshold $\tau$\\
\textbf{Output:} Retrained HQCNN model $M'$
\begin{algorithmic}[1]
\State Initialize: error counter $E$, generation proportion $R$, augmented sample set $D_{aug} = []$
\For{each $y \in D_{test}$}
    \State Predict $y'$ with $M$; if $y' \ne y$, then $E[y] \gets E[y] + 1$
\EndFor
\State $E_{total} \gets \sum_i E[i]$; for each class $i$, compute $R[i] \gets E[i] / E_{total}$
\State Adjust the QGAN training strategy based on error proportion $R$:
    \begin{enumerate}[a.]
        \item Increase weights of high-error classes $i$ in $G$
        \item Set training distribution of $D$ based on $R$
    \end{enumerate}
\State Train generator $G$ and discriminator $D$ to generate samples
\For{each class $i$}
    \State $N_i \gets N_{gen} \cdot R[i]$, generate $3N_i$ samples $S_i$
    \State Initialize filtered set $S_i' \gets []$
    \State Compute the specific confidence threshold $\tau$ for the class:
        \begin{enumerate}[a.]
            \item If $E[i]$ is large: $\tau = \tau - \alpha * R[i]$
            \item If $E[i]$ is small: $\tau = \tau + \beta * R[i]$
        \end{enumerate}
    \For{each sample $s$ in $S_{i}$}
        \State Predict with $M$ to get class $c$ and confidence $p$, if $c = i$ and $p \geq \tau$ add $s$ to $S_{i}'$ 
    \EndFor
    \While{$|S_{i}'| < N_i$}
        \State Regenerate until requirements met or max attempts reached 
    \EndWhile
    \State Select first $N_i$ samples from $S_{i}'$ and add to $D_{aug}$
\EndFor
\State Combine datasets $D_{combined} = D_{train} \cup D_{aug}$, Retrain model $M$ 
\State \textbf{Return} $M'$ 
\end{algorithmic}
\end{algorithm}

\section{Experimental Results and Analysis}\label{sec4}

This study conducts simulations on PyTorch and Qiskit using the MNIST dataset, a widely used benchmark in deep learning and quantum machine learning. MNIST consists of 28×28 grayscale digit images and serves as a standard benchmark for image classification \cite{bib43}. Due to the high computational cost of simulating large-scale quantum systems, the MNIST task is simplified to a three-class problem with digits "0", "1", and "2". This approach simplifies quantum circuit design and fits current hardware limitations. To simulate real-world data scarcity, only 100 samples per class are used for training and testing, significantly reducing data size. This setting allows in-depth evaluation of HQCNN data augmentation using QGANs under resource constraints. We next analyze the results of applying QGANs, classical GANs, and conventional data augmentation (CDA) to HQCNNs. By comparing training efficiency and generation quality under identical datasets and hyperparameters, we quantify QGAN’s advantages in parameter efficiency, iteration count, and resource usage. These findings offer theoretical and practical insights for advancing quantum data augmentation.

\subsection{Comparison of GANs with Different Architectures}\label{subsec4.1}

This section outlines the QGAN and classical GAN models used for comparison. The comparison includes the number of generator parameters and training iterations, as shown in Table \ref{table 1}.

\begin{table}[htbp]
    \centering
    \caption{Parameters and Training Epochs of Different GANs}
    \label{table 1}
    \begin{tabular}{lrr}
        \toprule
        \textbf{Model} & \textbf{Parameters} & \textbf{Epochs} \\
        \midrule
        QGAN   & 1,510,212 & 200 \\
        DCGAN  & 3,346,476 & 200 / 500 \\
        GAN-1  & 1,510,425 & 200 \\
        GAN-2  & 4,414,537 & 500 \\
        \bottomrule
    \end{tabular}
\end{table}

\subsection{Performance of the General-purpose Data Augmentation Strategy}\label{subsec1}

\begin{figure*}[htbp]
    \centering
    \begin{minipage}[t]{1\linewidth}
        \centering
        \includegraphics[width=5in]{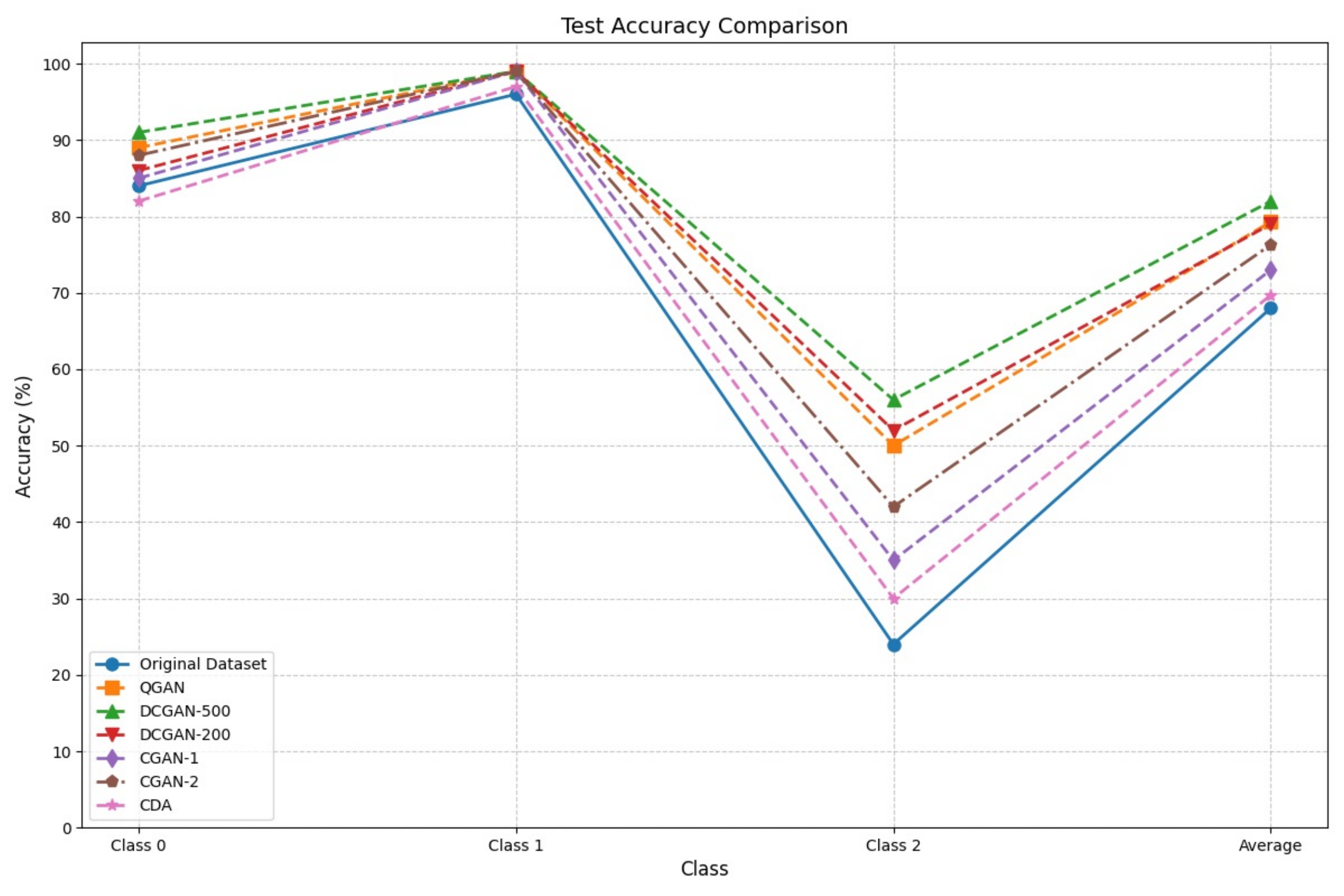}
    \end{minipage}
    \caption{Comparison of Different Data Augmentation Methods}
    \label{Fig.4}
\end{figure*}

An effective data augmentation method should be broadly applicable. Therefore, in this section, we evaluate the performance of QGAN and classic GANs of various architectures on HQCNN and compare them with traditional augmentation methods (random rotation, translation, and contrast adjustment) to assess the general strategy’s effectiveness. Following Algorithm \ref{Al.2}, the strategy generates 300 augmented samples, evenly distributing 100 per class across digits "0", "1", and "2" to ensure class balance. The detailed experimental comparison results are shown in Fig. \ref{Fig.4}.

An effective data augmentation method should be broadly applicable. 

The experimental results yield two key findings, offering empirical support for applying quantum generative adversarial networks (QGANs) to data augmentation in hybrid quantum-classical neural networks (HQCNNs).

First, the accuracy curve shows that GAN-based augmentation consistently outperforms conventional methods. However, contrary to Ref. \cite{bib44}, while conventional augmentation offers limited improvement for HQCNNs, it does not reduce classification accuracy. This may be because our HQCNNs use angle encoding and a hybrid quantum-classical architecture. This design helps preserve original features during quantum encoding, avoiding accuracy loss often caused by traditional augmentation. It also shows that the data diversity of traditional methods is limited, and it is difficult to overcome the performance bottleneck of HQCNN.

Secondly, QGAN demonstrated distinct advantages in the GAN-based augmentation comparison. Compared with GAN-1 (similar structure and parameters) and GAN-2 (more complex architecture), QGAN improved classification accuracy on MNIST by 6.3 and 3 percentage points, respectively, highlighting the superiority of quantum architectures in data generation. It is worth noting that QGAN achieves similar performance to DCGAN \cite{bib45} with much fewer parameters and iterations, balancing efficiency and effectiveness. Additionally, quantum entanglement may enhance sample quality and diversity by implicitly modeling complex data correlations.

In summary, through multiple controlled experiments, this study confirms the effectiveness of the QGAN framework in enhancing HQCNN accuracy and performance, offering theoretical and practical guidance for quantum data augmentation.

\subsection{Performance of the Customized Data Augmentation Strategy}\label{subsec1}

This section presents experimental results of data augmentation strategies tailored for hybrid quantum-classical neural networks (HQCNNs). We first report initial performance metrics. We then compare results across three aspects: dynamic vs. non-dynamic sample generation, the performance of different GAN variants, and the trade-off between sample quality and quantity. The performance indicators of the initial classifier are shown in Table \ref{table 2}.

\begin{table}[htbp]
    \centering
    \caption{Classification Accuracy and Average Prediction Confidence}
    \label{table 2}
    \begin{tabular}{lcccc}
        \toprule
        \textbf{Class} & \textbf{0} & \textbf{1} & \textbf{2} & \textbf{Average} \\
        \midrule
        Accuracy (\%) & 84 & 96 & 24 & 69 \\
        Average Prediction Confidence & 0.485 & 0.505 & 0.315 & 0.435 \\
        \bottomrule
    \end{tabular}
\end{table}

\subsubsection{Customized Augmentation \textbf{vs.} General Augmentation}\label{subsec1}

Based on Algorithm \ref{Al.3} and Table \ref{table 2}, we generated 300 samples. We then determined the number of samples assigned to each class. For well-performing categories, we retained images with confidence no lower than the initial value. These high-quality samples not only improved model performance for those classes but also helped balance the dataset. For underperforming categories, we lowered the threshold to include more samples, enhancing the model’s learning capacity. Details are provided in Table \ref{table 3}.  adversarial network (QGAN) as an example:	This experiment uses samples generated by the quantum generative adversarial network (QGAN) as an example:

\begin{table}[htbp]
    \centering
    \caption{Statistics for Different Classes}
    \label{table 3}
    \begin{tabular}{lccc}
        \toprule
        \textbf{Class} & \textbf{0} & \textbf{1} & \textbf{2} \\
        \midrule
        Number of Generated Samples & 45 & 11 & 244 \\
        Number of Augmented Dataset & 145 & 111 & 344 \\
        Confidence                  & 0.50 & 0.48 & 0.45 \\
        \bottomrule
    \end{tabular}
\end{table}

\begin{figure*}[htbp]
    \centering
    \begin{minipage}[t]{1\linewidth}
        \centering
        \includegraphics[width=5in]{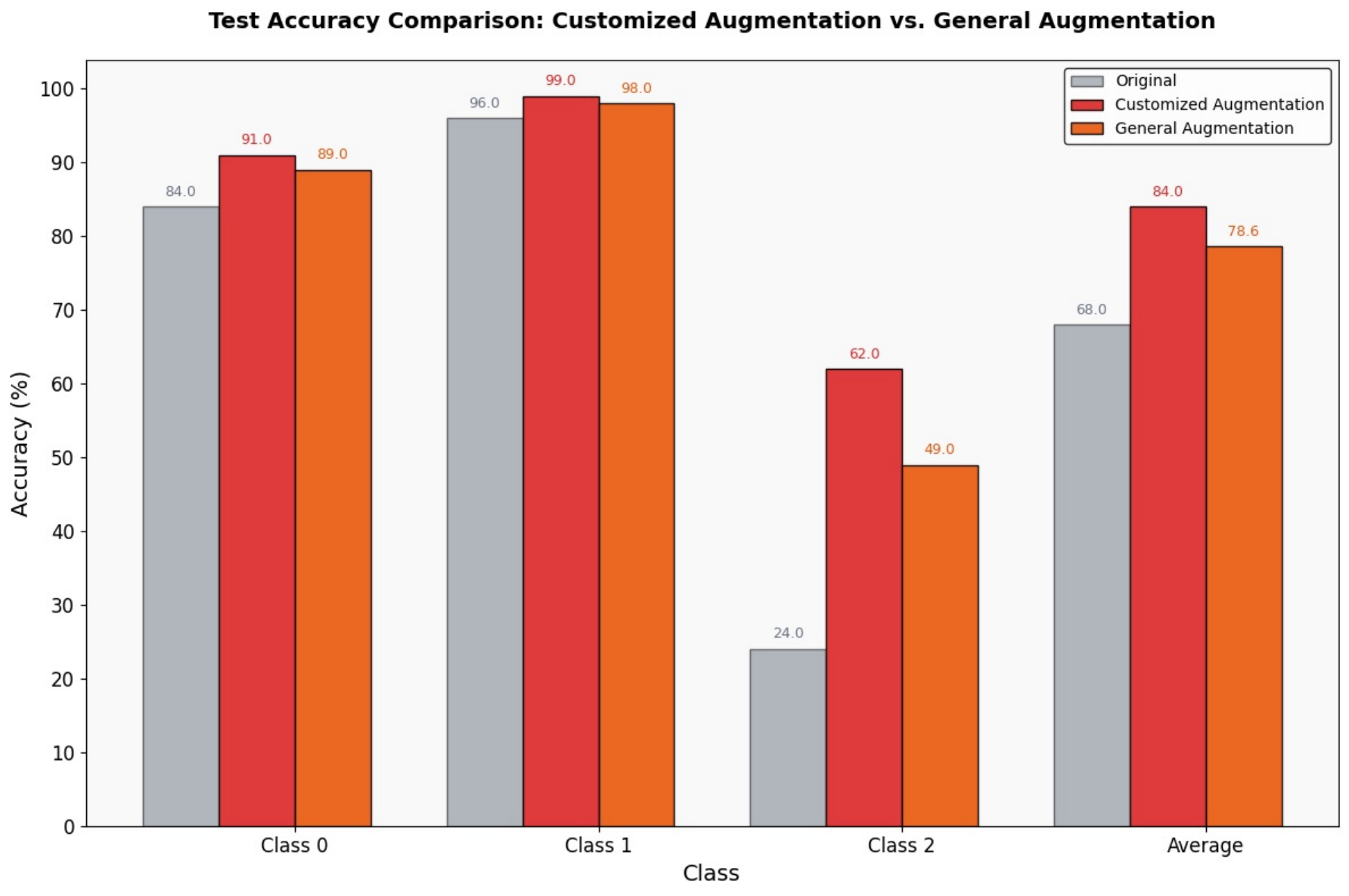}
    \end{minipage}
    \caption{Comparison between Customized Augmentation and General Augmentation}
    \label{Fig.5}
\end{figure*}

To establish a performance baseline, we implemented a general data augmentation strategy as a control. This strategy generates 300 augmented samples, with 100 per class to ensure balance. This strategy includes all generated samples in the training set without screening by confidence or class characteristics. This simulates the uniform expansion strategy often used in traditional augmentation, without considering the category-specific model performance. This control design highlights the value of the proposed customized strategy in boosting performance, while revealing the limitations of general augmentation in data-scarce settings. The comparison results are shown in Fig. \ref{Fig.5}.

\subsubsection{The Enhancement Effects of Various GANs}\label{subsec1}

\begin{figure*}[htbp]
    \centering
    \begin{minipage}[t]{1\linewidth}
        \centering
        \includegraphics[width=5in]{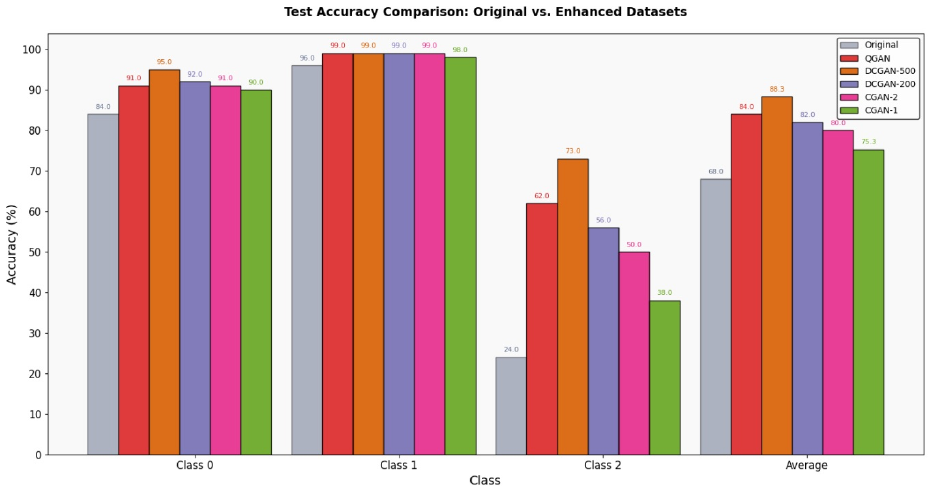}
    \end{minipage}
    \caption{Comparison of Various GANs Enhancement Effects}
    \label{Fig.6}
\end{figure*}

This study systematically compares the data augmentation effects of four GAN architectures (Sec. \ref{subsec4.1}) applied to HQCNNs. To ensure reliability and comparability, all four methods use a unified, customized augmentation strategy, applying dynamic sample generation and adjustable confidence screening thresholds for category-specific enhancement. The experiment controls key variables—including total generated data, sample quality, and HQCNN architecture—focusing on performance differences among classical GANs, DCGANs, and QGANs in enhancing HQCNN classification. Comparative results are shown in Fig. \ref{Fig.6}.

The experimental results show that the four types of GANs are effective in enhancing the performance of the specific HQCNN. Among them, DCGAN with 500 epochs achieved the best test accuracy (88.3\%), confirming its effectiveness in image generation and feature extraction. Notably, the quantum generative adversarial network (QGAN) significantly outperforms its classical counterpart. Under identical experimental conditions, QGAN reduces parameters by two-thirds and training iterations by 60\% compared to classical GAN-2, yet achieves higher classification accuracy. This highlights the unique strengths of quantum computing in data augmentation. Comparison with DCGAN further reveals QGAN’s technical advantages. Although QGAN slightly underperforms DCGAN at 500 iterations, it surpasses DCGAN at 200 iterations (84\% vs. 82\%) using a shallower network and half the parameters. These results suggest that QGAN performs efficiently under limited resources and holds promise as a substitute for traditional deep models in constrained environments (e.g., quantum hardware), offering both theoretical and practical value for quantum data augmentation. As quantum computing advances, QGAN is poised to drive innovation in machine learning.

\subsubsection{Sample Quality \textbf{vs.} Sample Quantity}\label{subsec1}

\begin{figure*}[htbp]
    \centering
    \begin{minipage}[t]{1\linewidth}
        \centering
        \includegraphics[width=5in]{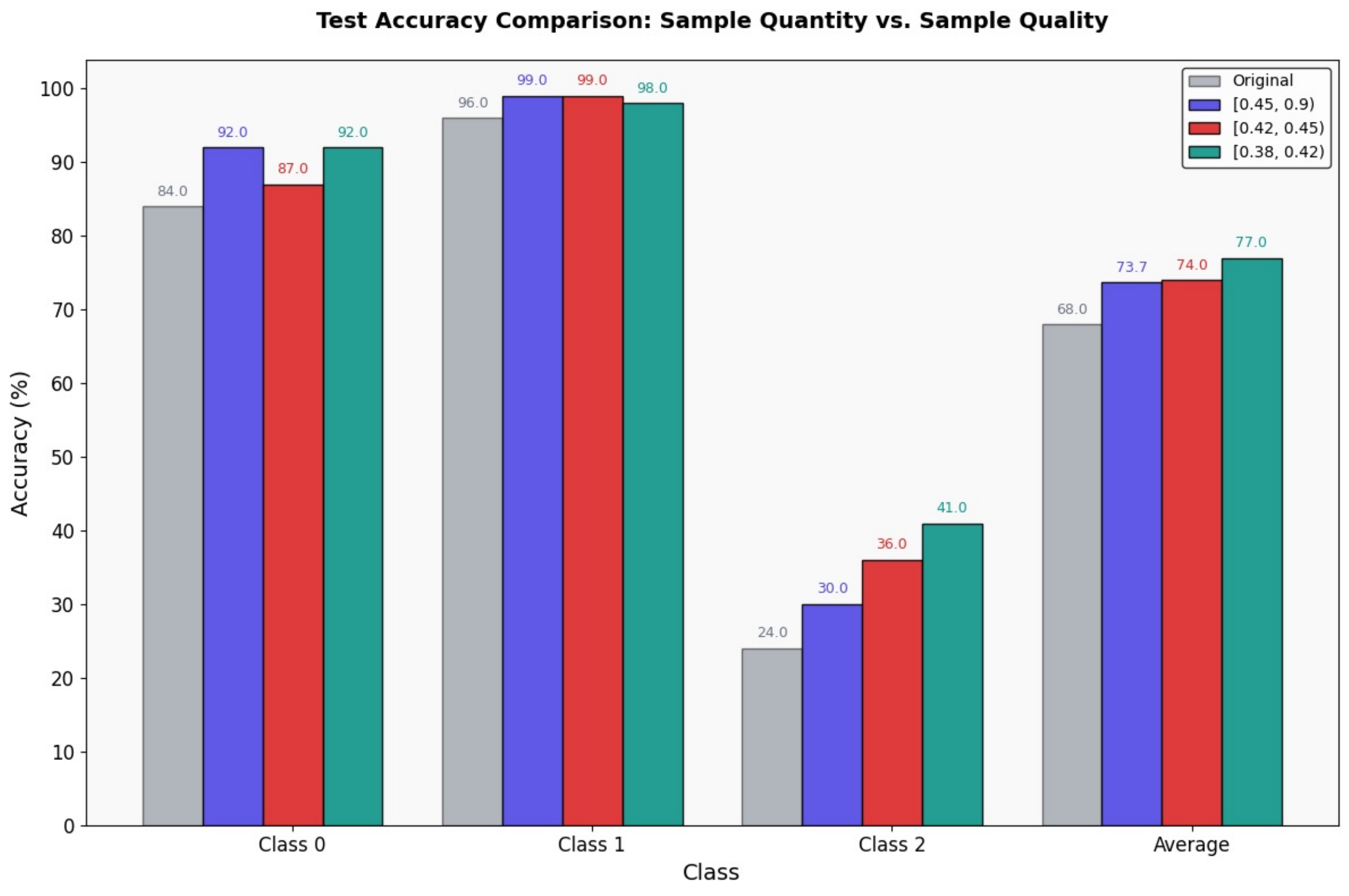}
    \end{minipage}
    \caption{Comparison of Sample Quality and Sample Quantity} 
    \label{Fig.7}
\end{figure*}

This study investigates how the quality and quantity of generated samples affect data augmentation. Using systematic experimental design, we conducted multiple controlled experiments to analyze HQCNN performance under different sample quality-quantity combinations. We controlled variables such as model architecture and training hyperparameters, and adjusted the confidence screening threshold to vary sample quality and quantity. Table \ref{table 4} details the experimental settings, and Fig. \ref{Fig.7} shows HQCNN classification accuracy under different sample quality-quantity conditions.

\newcolumntype{C}[1]{>{\centering\arraybackslash}p{#1}}  

\begin{table}[htbp]
    \centering
    \caption{Statistics of Generated Samples and Confidence}
    \label{table 4}
    \begin{tabular}{lccccc}
        \toprule
        \textbf{Class} & \textbf{0} & \textbf{1} & \multicolumn{3}{c}{\textbf{2}}\\
        \midrule
        Generated Samples & 45 & 11 & 9 & 16 & 43 \\
        Augmented Dataset & 145 & 111 & 109 & 116 & 143 \\
        Confidence  & 0.50 & 0.48 & $[0.45,\ 0.9)$ & $[0.42,\ 0.45)$ & $[0.38,\ 0.42)$ \\
        \bottomrule
    \end{tabular}
\end{table}

Experimental results show that under constraints on sample quantity and generator performance, increasing the number of augmented samples in underperforming categories is more effective than improving sample quality. This suggests that in resource-constrained settings, prioritizing sample quantity in key categories more efficiently addresses model shortcomings, offering practical guidance for optimizing data augmentation strategies.

\section{Conclusion}\label{sec5}

To tackle data scarcity in quantum machine learning (QML), we propose a data augmentation framework that integrates hybrid quantum-classical neural networks (HQCNNs) with quantum generative adversarial networks (QGANs). The framework includes two complementary strategies: a general strategy adaptable to various HQCNN architectures, and a customized strategy tailored to specific model weaknesses. Together, these strategies substantially improve model performance. Experiments on the MNIST dataset demonstrate the framework’s effectiveness. Compared to traditional data augmentation techniques and classical GAN models, the QGAN-based augmentation strategy exhibits exceptional performance. Remarkably, with far fewer parameters than deep convolutional GANs (DCGANs), our approach achieves similar results, highlighting quantum computing’s efficiency in generative modeling. These findings break through the limitations of conventional data augmentation methods, paving a transformative path for advancements in QML.

Future work should enhance QGAN’s ability to generate diverse, high-quality samples at scale to meet complex task demands. Additionally, deeper integration of HQCNNs and QGANs is also needed to improve efficiency and accuracy in high-dimensional data processing. Progress in quantum hardware will be key to bridging the gap between theory and practice, enabling real-world deployment of quantum-enhanced machine learning.



\section*{Declaration of competing interest}
The authors declare that they have no known competing financial interests or personal relationships that could have appeared
to influence the work reported in this paper.

\section*{Acknowledgement}
This work is supported by National Natural Science Foundation of China (Grant Nos. 62371069, 62372048,  62272056).

\section*{Data availability}
No data was used for the research described in the article.





\end{document}